\documentclass[sn-nature]{bst/sn-jnl}
\jyear{2023}
\usepackage{amsmath,amssymb,amsfonts}
\usepackage{tabulary,fancyhdr,amsbsy,latexsym}
\usepackage{url,morefloats,floatflt,cancel,tfrupee}
\usepackage{colortbl}
\usepackage{pifont}
\usepackage[nointegrals]{wasysym}
\usepackage{float}
\usepackage{siunitx}
\usepackage{tabularx, booktabs}
\usepackage{lipsum}
\usepackage{fix-cm}
\newcommand{\figurepath}[1]{figures/#1}
\definecolor{mygreen}{HTML}{E9ECE6}
\usepackage{framed}

\begin{document}

\title[Article Title]{Crowd-Certain: Label Aggregation in Crowdsourced and Ensemble Learning Classification}
\author[]{\fnm{Mohammad S.} \sur{Majdi}}
\author[]{\fnm{Jeffrey J.} \sur{Rodriguez}}
\affil[]{\orgdiv{Dept of Electrical and Computer Engineering}, \orgname{The University of Arizona}, \orgaddress{\city{Tucson}, \state{AZ}, \country{USA}}}

\abstract{
        Crowdsourcing systems have been used to accumulate massive amounts of labeled data for applications such as computer vision and natural language processing.
        However, because crowdsourced labeling is inherently dynamic and uncertain, developing a technique that can work in most situations is extremely challenging.
        In this paper, we introduce Crowd-Certain, a novel approach for label aggregation in crowdsourced and ensemble learning classification tasks that offers improved performance and computational efficiency for different numbers of annotators and a variety of datasets.
        The proposed method uses the consistency of the annotators versus a trained classifier to determine a reliability score for each annotator.
        Furthermore, Crowd-Certain leverages predicted probabilities, enabling the reuse of trained classifiers on future sample data, thereby eliminating the need for recurrent simulation processes inherent in existing methods.
        We extensively evaluated our approach against ten existing techniques across ten different datasets, each labeled by varying numbers of annotators.
        The findings demonstrate that Crowd-Certain outperforms the existing methods (Tao, Sheng, KOS, MACE, MajorityVote, MMSR, Wawa, Zero-Based Skill, GLAD, and Dawid Skene), in nearly all scenarios, delivering higher average accuracy, F1 scores, and AUC rates.

        Additionally, we introduce a variation of two existing confidence score measurement techniques. Finally we evaluate these two confidence score techniques using two evaluation metrics: Expected Calibration Error (ECE) and Brier Score Loss.
        Our results show that Crowd-Certain achieves higher Brier Score, and lower ECE across the majority of the examined datasets, suggesting better calibrated results.
    }

\maketitle
\section{Introduction}\label{sec:crowd.Introduction}
Supervised learning techniques require a large amount of labeled data to train models to classify new data~\cite{jiang_Wrapper_2019,jiang_Class_2019}. Traditionally, data labeling has been assigned to experts in the domain or well-trained workers~\cite{tian_MaxMargin_2019}. Although this method produces high-quality labels, it is inefficient and costly~\cite{li_Noise_2016,li_Noise_2019}.  Social networking provides an innovative solution to the labeling problem by allowing data to be labeled by online crowd workers (annotators). This has become feasible, as crowdsourcing services such as Amazon Mechanical Turk (formerly CrowdFlower) have grown in popularity. Crowdsourcing systems have been used to accumulate large amounts of labeled data for applications such as computer vision~\cite{deng_ImageNet_2009,liu_Variational_2012} and natural language processing~\cite{karger_Budget_2014}. However, because of individual differences in preferences and cognitive abilities, the quality of labels acquired by a single crowd worker is typically low, thus jeopardizing applications that rely on these data. This is because crowd workers are not necessarily domain experts and may lack the necessary training or expertise to produce high-quality labels.

Aggregation after repeated labeling is one method for handling workers with various abilities. Label aggregation is a process used to infer an aggregated label for a data instance from a multi-label set~\cite{sheshadri_SQUARE_2013}. Several studies have demonstrated the efficacy of repeated labeling~\cite{tu_MultiLabel_2018,zhang_multilabelinferencecrowdsourcing_2018}. Repeat labeling is a technique in which the same data are labeled by multiple workers, and the results are combined to estimate an aggregated label using majority voting (MV) or other techniques. In the case of MV, an aggregated label is the label that receives the most votes from the workers for a given data instance. This can help reduce the impact of biases or inconsistencies made by workers. Several factors, such as problem-specific characteristics, the quality of the labels created by the workers, and the amount of data available, can influence the effectiveness of the aggregation methodologies. Consequently, it is difficult to identify a clear winner among the different techniques. For example, in binary labeling, one study~\cite{sheshadri_SQUARE_2013} discovered that Raykar's~\cite{raykar_Learning_2010} technique outperformed other aggregation techniques. However, according to another study~\cite{zheng_Truth_2017}, the traditional Dawid-Skene (DS) model~\cite{dawid_Maximum_1979} was more reliable in multi-class settings (where data instances can be labeled as belonging to multiple classes).

Furthermore, regardless of the aggregation technique used, the performance of many aggregation techniques in real-world datasets remains unsatisfactory~\cite{liu_Exploiting_2021}. This can be attributed to the complexity of these datasets, which often do not align with the assumptions and limitations of different methods. For example, real-world datasets may present issues such as labeling inaccuracies, class imbalances, or overwhelming sizes that challenge efficient processing with available resources. These factors can adversely affect the effectiveness of label aggregation techniques, potentially yielding less than optimal results for real-world datasets.
Prior information may be used to enhance the label aggregation procedure.

This can include domain knowledge, the use of quality control measures, and techniques that account for the unique characteristics of workers and data. Knowing the reliability of certain workers, it is possible to draw more accurate conclusions about labels~\cite{li_Crowdsourced_2017}. For instance, in the label aggregation process, labels produced by more reliable workers (such as domain experts) may be given greater weight. The results of the label aggregation process can also be validated using expert input~\cite{liu_Improving_2017}. During the labeling process, domain experts can provide valuable guidance and oversight to ensure that the labels produced are accurate and consistent.
The agnostic requirement for general-purpose label aggregation is that label aggregation cannot use information outside the labels themselves. This requirement is not satisfied in most label aggregation techniques~\cite{zhang_Crowdsourced_2019}. The agnostic requirement ensures that the label aggregation technique is as general as possible and applicable to a wide range of domains with minimal or no additional context.

The uncertainty of workers during labeling can provide valuable prior knowledge to determine the appropriate amount of confidence to grant each worker while still adhering to the requirement of a general-purpose label aggregation technique. We developed a method for estimating the reliability of different workers based on the worker's own consistency during labeling. We take this concept a step further by calculating a weight for each worker based not only on their reliability but also on their agreements with other workers involved. This consideration of inter-reliability ensures a more comprehensive and dynamic weighting process, adjusting to the overall performance of the entire group of workers. Thus, the generated weights become a measure of both individual and collective reliability, which significantly improves the accuracy of our labeling aggregation method.

We introduce a novel method, named Crowd-Certain, which offers a significant improvement in label aggregation for crowdsourced and ensemble learning classification tasks, yielding improved performance across various scenarios. This technique leverages the consistency of workers versus a trained classifier to ascertain their reliability, resulting in more accurate and efficient label aggregation.

Our extensive experimental evaluation, conducted across ten diverse datasets, demonstrates that Crowd-Certain outperforms established techniques (Gold Majority Vote, MV, MMSR, Wawa, Zero-Based Skill, GLAD, and Dawid Skene) in terms of aggregated label accuracy compared to ground truth labels. Importantly, Crowd-Certain generates weights that closely follow pre-set ground truth accuracy for each worker (referred to as a probability threshold in this study). Moreover, during inference time, Crowd-Certain employs predicted probabilities (obtained from a classifier trained on the worker's label set) rather than worker's labels, which facilitates the reuse of trained classifiers for future data samples, removing the need for repeating the training process when new data samples are introduced.
The remainder of this paper is organized as follows. Section~\ref{sec:crowd.relatedwork} examines related work involving label aggregation algorithms. In Section~\ref{sec:crowd.method}, we provide an in-depth explanation of Crowd-Certain. Section~\ref{sec:crowd.results} presents the experiments and findings, and Section~\ref{sec:crowd.discussion} encapsulates the results, highlighting key insights. Lastly, Section~\ref{sec:crowd.conclusion} concludes the paper and highlights the potential directions for future research.

\section{Related Work}\label{sec:crowd.relatedwork}
Numerous label aggregation algorithms have been developed to capture the complexity of crowdsourced labeling systems, including techniques based on worker reliability~\cite{bi_Learning_2014,demartini_Zencrowd_2012}, confusion matrices~\cite{raykar_Learning_2010,zhang_Spectral_2014}, intentions~\cite{bi_Learning_2014,kurve_MultiCategory_2015}, biases~\cite{zhang_Imbalanced_2013,hernandez-gonzalez_Note_2019, welinder_Multidimensional_2010}, and correlations~\cite{ma_Gradient_2020}. However, because crowdsourced labeling is inherently dynamic and uncertain, developing a technique that can work in most situations is extremely challenging. Many techniques~\cite{liu_Variational_2012,karger_Budget_2014,raykar_Learning_2010,dalvi_Aggregating_2013,ghosh_Who_2011} utilize the Dawid and Skene (DS) generative model~\cite{dawid_Maximum_1979}. Ghosh~\cite{ghosh_Who_2011} extended the DS model by using singular value decomposition (SVD) to calculate the reliability of the worker. Similarly, to Ghosh~\cite{ghosh_Who_2011}, Dalvi~\cite{dalvi_Aggregating_2013} used SVD to estimate true labels with a focus on the sparsity of the labeling matrix. In crowdsourcing, it is common for the labeling matrix to be sparse, meaning that not all workers have labeled all the data. This may be due to several factors, such as the cost of labeling all data instances or the workers' time constraints. Karger~\cite{karger_Budget_2014} described an iterative strategy for binary labeling based on a one-coin model~\cite{ghosh_Who_2011}. Karger~\cite{karger_Budget_2014} extends the one-coin model to multi-class labeling by converting the problem into $k-1 $ binary problems (solved iteratively), where $k $ is the number of classes.

The MV technique assumes that all workers are equally reliable. For segmentation, Warfield~\cite{warfield_Simultaneous_2004} proposed simultaneous truth and performance level estimation (STAPLE), a label fusion method based on expectation maximization. STAPLE ``weighs'' expert opinions during label aggregation by modeling their reliability. Since then, many variants of this technique have been proposed~\cite{winzeck_ISLES_2018,commowick_Objective_2018,asman_Robust_2011,asman_Formulating_2012, eugenioiglesias_Unified_2013, jorgecardoso_STEPS_2013,asman_NonLocal_2013,akhondi-asl_Logarithmic_2014}. The problem with these label aggregation approaches is that they require the computation of a unique set of weights for each sample, necessitating the re-evaluation of the workers' weights when a new instance is added.

Among the numerous existing label aggregation strategies, MV remains the most efficient and widely used approach~\cite{tao_Label_2020}. If we assume that all workers are equally reliable and that their errors are independent of one another, then, according to the theory of large numbers, the likelihood that the MV is accurate increases as the number of workers increases. However, the assumption that all workers are equally competent and independent may not always hold. Furthermore, MV does not provide any additional information on the degree of disagreement among the workers (As an example, consider the scenario where four of seven doctors think patient A needs immediate surgery, while all seven think patient B needs immediate surgery; MV will simply label ``yes'' in both cases).
To address this problem, additional measures such as inter-worker agreement (IAA) have been used~\cite{artstein_InterAnnotator_2017}. IAA is a measurement of the agreement among multiple workers who label the same data instance. Typically, IAA is calculated using statistical measures, such as Cohen's kappa, Fleiss's kappa, or Krippendorff's alpha~\cite{krippendorff_Content_2018}. These measures consider both the observed agreement between the workers and the expected agreement owing to random chance. IAA can also be visualized using a confusion matrix or annotation heatmap, which illustrates the distribution of labels assigned by the workers. This can help identify instances where the workers disagree or are uncertain and can guide further analysis to improve the annotation~\cite{carletta_Assessing_1996}.

Recently, Sheng~\cite{sheng_Majority_2019} proposed a technique that provided a confidence score along with an aggregated label. The main problem with this approach is that it assumes that all workers are equally capable when calculating the confidence score. Tao~\cite{tao_Label_2020} improved Sheng's approach by assigning different weights to workers for each instance. This weighting method combines the specific quality $s_\alpha^{(i,k)} $ for the worker $\alpha $ and instance $i $ and the overall quality $\tau_\alpha$ across all instances.

Inspired by Li's technique~\cite{li_Incorporating_2018}, Tao evaluates the similarity between the worker labels for each instance. To derive the specific quality $s_{\alpha}^{(i)}$, Tao counts the number of workers who assigned the same label as the worker $\alpha $ for that instance. To calculate the overall quality $\tau_\alpha $, Tao performs a 10-fold cross-validation to train each of the 10 classifiers on a different subset of data using the labels provided by the worker $\alpha $ as true labels and then assigns the average accuracy of the classifiers across all remaining instances as $\tau_\alpha $. The final weight for worker $\alpha $ and instance $i $ is then calculated using the sigmoid function $\gamma_{i,\alpha}=\tau_\alpha\left(1+{\left(s_{\alpha}^{(i)}\right)}^{2}\right) $. However, Tao's technique~\cite{tao_Label_2020} has some drawbacks. It relies on the labels of other workers to estimate $s_{\alpha}^{(i)} $. However, different workers have varying levels of competence (reliability) when labeling the data, and therefore, relying on their labels to measure $s_{\alpha}^{(i)} $ will result in propagating the errors and biases of their labels during weight estimation. Furthermore, Tao's technique~\cite{tao_Label_2020} relies on the labels provided by each worker $\alpha $ to estimate their respective $\tau_\alpha $ by assuming that the trained classifiers can learn the inherent characteristics of the datasets even in the absence of ground truth labels. While that may be true in some cases, it typically leads to suboptimal measurement and the propagation of biases and errors, from both the worker's labels and the classifier, into weight estimation.

\section{Methods}\label{sec:crowd.method}
We propose a novel method called Crowd-Certain which focuses on leveraging uncertainty measurements to improve decision-making in crowdsourcing and ensemble learning scenarios. Crowd-Certain employs a weighted soft majority voting approach, where the weights are determined based on the uncertainty associated with each worker's labels. Initially, we use uncertainty measurement techniques to calculate the degree of consistency of each worker during labeling.

Furthermore, to ensure that the proposed technique does not calculate a high weight for workers who are consistently wrong (for example, when a specific worker always mislabels a specific class, and hence demonstrates a high consistency even if they label instances incorrectly), we extend the proposed technique by penalizing the workers for instances in which they disagree with the aggregated label obtained using MV. To mitigate the reliance on training a classifier on an worker's labels, which may be inaccurate, we train an ensemble of classifiers for each worker. In addition, we report two confidence scores along with the aggregated label to provide additional context for each calculated aggregate label. We report a single weight for all instances in the dataset. As will be demonstrated in Section~\ref{sec:crowd.results}, the proposed Crowd-Certain method is not only comparable to other techniques in terms of accuracy of the aggregated labels with respect to the ground truth labels for scenarios with a large number of workers, but also provides a significant improvement in accuracy for scenarios where the number of workers may be limited. Furthermore, by assigning a single weight to each worker for all instances in the dataset, the model can assign labels to new test instances without recalculating the worker weights. This is especially advantageous in situations where workers are scarce as it enables the model to make accurate predictions with minimal dependence on the worker input. This characteristic of the Crowd-Certain method can significantly reduce the time and resources required for labeling in practical applications. When deploying the model in real-world scenarios such as medical diagnosis, fraud detection, or sentiment analysis, it could be advantageous to be able to assign labels to new instances without constantly recalculating worker weights.

\subsection{Glossary of Symbols}
For convenience, the following list summarizes the major symbols used in the subsequent discussion:
\begin{itemize}{
    \setlength{\parindent}{1.5em}
    \item  $N$: Number of instances.
    \item  $M$: Number of workers.
    \item  $y^{(i,k)} \in \{0,1\} $: True label for the $k $-th class for instance $i $.
    \item  $z_\alpha^{(i,k)} \in \{0,1\} $: Label given by worker $\alpha $ for $k $-th class for instance $i $.
    \item  ${{\underset\alpha{\mathrm{MV}}}{\left(z_\alpha^{(i,k)}\right)}} $: Majority voting technique (the label that receives the most votes) applied to worker labels for class $k $ and instance $i $.
    \item  $\pi_\alpha^{(k)} $: Probability threshold used as a pre-set ground truth accuracy, for each worker $\alpha$ and class $k$. It is used to generate sample binary labels (fictitious ground truth label set) for worker $\alpha $ for class $k $. For example, the threshold values may be obtained from a uniform distribution in the interval $0.4 $ to $1 $, i.e., $\pi_\alpha^{(k)} \sim U(0.4,1) $.
    \item  $X^{(i)} $: Data for instance $i$.
    \item  $Y^{(i)}=\left\{y^{(i,1)},y^{(i,2)},\;\dots,y^{(i,K)}\right\} $: True label set, for instance $i $. For example, consider a dataset that is labeled for the presence of cats, dogs, and rabbits in any given instance. If a given instance $X^{(i)} $ has cats and dogs but not rabbits, then $Y^{(i)}=\{1,1,0\} $.
    \item  $Z_{\alpha}^{(i)}=\left\{z_\alpha^{(i,1)}, z_\alpha^{(i,2)}, \dots, z_\alpha^{(i,K)}\right\} $: Label set given by the worker $\alpha $ for instance $i $.
    \item $K$: number of categories (aka classes) in a multi-class multi-label problem. For example, if we have a dataset labeled for the presence of cats, dogs, and rabbits in any given instance, then $K=3$.
    \item  $\rho^{(i)}$: Randomly generated number  between 0 and 1 for instance $i $. It is obtained from a uniform distribution, i.e., $\rho^{(i)} \sim U(0,1) $ This number is used to determine, for each instance $i$, whether the true label should be assigned to each fictitious worker's label. For each class $k$, if the worker's probability threshold $\pi_\alpha^{(k)}$ is greater than $\rho^{(i)}$, the true label $y^{(i,k)}$ is assigned; otherwise, an incorrect label $1 - y^{(i,k)}$  is assigned.
    \item  $\Pi_\alpha=\left\{ \pi_\alpha^{(1)} , \pi_\alpha^{(2)}  , \dots, \pi_\alpha^{(K)} \right\} $: set of $K $ probability thresholds for worker $\alpha $.
    \item  $\mathbb{X}={\left\{X^{(i)}\right\}}_{i=1}^{N} $: Set of all instances.
    \item  $\mathbb{Y} = {\left\{Y^{(i)}\right\}}_{i=1}^{N} $: Set of all true labels.
    \item  $\mathbb{Z}_\alpha = {\left\{Z_\alpha^{(i)}\right\}}_{i=1}^{N} $: Set of all labels for the worker $\alpha $.
    \item  $\mathbb{P} = {\left\{\rho^{(i)}\right\}}_{i=1}^{N} $: Set of $N $ randomly generated numbers.
    \item  $\mathbb{D}=\left\{\mathbb{X},\mathbb{Y}\right\} $: Dataset containing all instances and all true labels.
    \item  $\mathbb{D}_\alpha=\left\{\mathbb{X},\mathbb{Z}_\alpha\right\} $: Dataset containing the labels given by the worker $\alpha $.
    \item  $f_{\alpha}^{(g)}(\cdot)$: Classifier $g $ trained on dataset $\mathbb{D}_{\alpha}^{\mathrm{train}} $ with random seed number $g $ (which is also the classifier index)
    \item  $P_{\alpha}^{(i),(g)} = {\left\{ p_{\alpha}^{(i,k),(g)} \right\}}_{k=1}^{K} $: Predicted probability set obtained in the output of the classifier $f_{\alpha}^{(g)}(\cdot) $ representing the probability that each class $k $ is present in the sample.
    \item  $\theta_{\alpha}^{(k),(g)} $: Binarization threshold. To obtain this, we can utilize any existing thresholding technique. For example, in one technique, we analyze the ROC curve and find the corresponding threshold where the difference between the true positive rate (sensitivity) and false positive rate (1-specificity) is maximum. Alternatively, we could simply use $0.5 $.
    \item  $t_\alpha^{(i,k),(g)} =
        \begin{cases}
            1 & \text{if } p_\alpha^{(i,k),(g)} > \theta_\alpha^{(k),(g)}, \\
            0 & \text{otherwise}.
        \end{cases} $: Predicted label obtained by binarizing $p_\alpha^{(i,k),(g)} $.
    \item  $\eta_{\alpha}^{(i,k)} = {{\underset g{\mathrm{MV}}}{ \left(t_\alpha^{(i,k),(g)}\right) }} $: The output of the majority vote applied to the predicted labels obtained by the $G $ classifiers.
    \item  $\Delta_{\alpha}^{(i,k)} $: Uncertainty score.
    \item  $c_\alpha^{(i,k)} $: Consistency score.
    \item  $\omega_\alpha^{(k)} $: Estimated weight for worker $\alpha $ and class $k $.
    \item  $\nu^{(i,k)} = \frac{1}{{M}}{\sum_{\alpha}{\omega_\alpha^{(k)} \; \eta_\alpha^{(i,k)}}}$: Final aggregated label for class $k $ and instance $i $.
    }
\end{itemize}
\subsection{Risk Calculation}
Label aggregation is frequently used in various machine learning tasks, such as classification and regression, when multiple workers assign labels to the same data points. The aggregation model refers to the underlying function that maps a set of multiple labels, obtained by different workers, into one aggregated label. In the context of label aggregation, this model can be a neural network, a decision tree, or any other machine learning algorithm capable of learning to aggregate labels provided by multiple workers. The objective of this study is to develop an aggregation model capable of accurately determining true labels despite potential disagreements among workers. One common method to achieve this involves minimizing the total error (or disagreement) between the workers' assigned labels and the true labels, as follows:
\begin{equation}
    E = \sum_{i=1}^N \sum_{a=1}^M \left( \sum_{k=1}^K \delta\left(y^{(i,k)}, z_\alpha^{(i,k)}\right) \right)
    \label{eq:crowd.Eq.1.risk.error}
\end{equation}
where $\delta $ is the Kronecker delta function.

Although error is a crucial aspect in determining the aggregation model's performance, it treats false positives and false negatives with equal weight. However, in many practical scenarios, it is essential to weigh false positives and false negatives differently depending on the specific context and potential consequences of each type of misclassification. The concept of risk allows us to achieve this by incorporating a loss function, which assigns different weights to different types of errors. In this way, risk serves as a weighted calculation of error, enabling us to better evaluate the performance of an aggregation model and its generalization capability.

Let us denote loss function, $\mathcal{L}(\cdot)$, as a function that quantifies the discrepancy between the predicted labels and the true labels, accounting for the varying importance of different types of errors. Risk, denoted as $R(h) $, represents the expected value of a loss function over all possible data instances.  In practice, our goal is to minimize the risk to achieve optimal performance on unseen data. However, since we only have access to a limited dataset (empirical distribution), we instead work with the empirical risk. This limitation may arise because of the need to reserve a portion of our data for testing and validation or because no dataset can fully capture all possible data instances in the real world. However, minimizing risk alone could result in overfitting, in which the aggregation model learns the noise in the training data rather than the underlying patterns, resulting in poor generalization to unseen data. To improve generalizability, it is necessary to employ regularization techniques to strike a balance between the complexity of the aggregation model and its ability to fit the training data.

Risk measurement enables us to assess the aggregation model's performance in terms of accuracy (of the aggregated labels with respect to the ground truth labels), overfitting (when risk is minimized, but the model performs poorly on unseen data), and model complexity. Assume that the aggregation model $h (\cdot) $ is a function that takes a set of $M $ label sets $Z^{(i)} $ for each instance $i $ in the training data and calculates an aggregated label set $\widehat{Y}^{(i)} $ as an estimate of the true label set $Y^{(i)} $. Our goal is to find an aggregation model $h(\cdot) $ that minimizes risk defined as follows:
\begin{equation}
    R(h) = \frac{1}{N} \sum_{i=1}^{N} \mathcal{L} \left( Y^{(i)}, h\left({\left\{Z_{\alpha}^{(i)}\right\}}_{\alpha=1}^{M}\right)\right)
    \label{eq:crowd.Eq.2.risk.emp}
\end{equation}
In this context, $\mathcal{L}(\cdot) $ represents an arbitrary loss function, which quantifies the discrepancy between predicted labels and true labels while accounting for the varying importance of different types of errors.
Our goal is to choose an aggregation model $\widehat{h} $ that minimizes the risk, following the principle of risk minimization~\cite{vapnik_Principles_1991}:
\begin{equation}
    \widehat{h} = \underset{h}{\text{argmin}}  R(h)
    \label{eq:crowd.Eq.3.risk.h}
\end{equation}
\subsection{Generating Workers' Label Sets from Ground Truth}\label{subsec:methods.generating_fictitious_labelset}
In order to evaluate the proposed Crowd-Certain technique (with and without penalization) as well as other aggregation techniques, we create $M$ fictitious workers. To synthesize a multi-worker dataset from a dataset with existing ground truth, we use a uniform distribution in the interval from $0.4 $ to $1 $, i.e., $\pi_\alpha^{(k)} \sim U\left(0.4,1\right) $ (however other ranges can also be used) to obtain $M \times  K$ probability thresholds $\Pi $, where $K$ is the number of classes. Note that a worker may be skilled at labeling dogs, but not rabbits. Then we use these probability thresholds to generate the crowd label set $Z_{\alpha}^{(i)} $ from the ground truth labels for each instance $i $.

For each worker $\alpha $, each instance $i $ and class $k $ in the dataset is assigned its true label with probability $\pi_\alpha^{(k)}$ and the opposite label with probability $ (1-\pi_\alpha^{(k)})$. To generate the labels for each worker $\alpha $, a random number $0 < \rho^{(i)} < 1 $ is generated for each instance $i $ in the dataset. Then $\forall \alpha,k \; \; \text{if} \; \; \rho^{(i)}\leq \pi_\alpha^{(k)}$. Then the true label is used for that instance and class for the worker $\alpha $; otherwise, the incorrect label is used.
The calculated worker labels $z_{\alpha}^{(i,k)} $ for each worker $\alpha $, instance $i $ and class $k $ are as follows:
\begin{equation}
    z_{\alpha}^{(i,k)} =
    \begin{cases}
        y^{(i,k)} & \text{if} \rho^{(i)}  \leq \pi_\alpha^{(k)} , \\
        1 - y^{(i,k)} & \text{if} \rho^{(i)} > \pi_\alpha^{(k)} ,
    \end{cases} \quad \forall i, a, k
    \label{eq:crowd.Eq.4.fictitious_label}
\end{equation}
To evaluate the proposed techniques over all data instances, a k-fold cross-validation is employed.
\subsection{Uncertainty Measurement}\label{subsec:crowd.uncertainty}
A common approach to measure uncertainty is to increase the number of data instances $X $ in the test dataset $\mathbb{D}_\alpha^{\mathrm{test}} $ to create multiple variations of each sample data $X^{(i)} $~\cite{ayhan_TestTime_2018}. In this approach, for each instance $i $, we apply randomly generated spatial transformations and additive noise to the input data $X^{(i)} $ to obtain a transformed sample and repeat this process $G $ times to obtain a set of $G $ transformed samples.

However, this approach is mostly suitable for cases where the input data comprises images or volume slices. Since the datasets used in this study consist of feature vectors instead of images or volume slices, this approach cannot be used. To address this problem, we introduced a modified uncertainty measurement approach, in which instead of augmenting the data instances $X^{(i)} $, we feed the same sample data to different classifiers.
The steps are as follows.
\begin{itemize}
    \item For the choice of classifier, we can either use a probability-based classifier such as random forest and train it under $G $ different random states or train various classifiers and address the problem in a manner similar to ensemble learning (using a set of $G $ different classification techniques such as random forest, SVM, CNN, Adaboost, etc.~\cite{zhou_Ensemblelearning_2009}).
    \item In either case, we obtain a set of $G $ classifiers ${\left\{f_{\alpha}^{(g)}( \cdot)\right\}}_{g=1}^G $ for each worker $\alpha $. The classifier $f_{\alpha}^{(g)}( \cdot) $ is trained on a labeled training dataset $\mathbb{D}_\alpha^{\mathrm{train}}$. This training process enables $f_{\alpha}^{(g)}(\cdot) $ to learn the underlying patterns in the data and make predictions on unseen instances.
    \item The index value $g  \in \{1,2,\dots,G\} $ is used as the random seed value during training of the $g$\-th classifier for all workers.
    \item After training, we feed the test samples $X^{(i)}\in \mathbb{X}^{\text{test}} $ to the $g $-th classifier $f_{\alpha}^{(g)}(\cdot) $ as test cases.
    \item The classifier $f_{\alpha}^{(g)}(\cdot) $ then outputs a set of predicted probabilities ${\left\{p_{\alpha}^{(i,k),(g)}\right\}}_{k=1}^{K} $ representing the probability that class $k $ is present in the sample. Consequently, we obtain a collection of $G $ predicted probability sets ${\left\{ {\left\{ p_{\alpha}^{(i,k),(g)}\right\}}_{k=1}^K \right\}}_{g=1}^G $ for each worker $\alpha $ and instance $i $.
    \item The set ${\left\{p_{\alpha}^{(i,k),(g)}\right\}}_{g=1}^G $ contains the predicted probabilities for class $k $,
    worker $\alpha $, and instance $i $.
    \item Disagreements between predicted probabilities ${\left\{p_{\alpha}^{(i,k),(g)}\right\}}_{g=1}^G $ can be used to estimate uncertainty.
    \item The reason for using classifiers rather than using the crowdsourced labels directly is two-fold.
    \begin{enumerate}
        \item Using a probabilistic classifier helps us calculate uncertainty based on each worker's labeling patterns that the classifier learns.
        \item Furthermore, this approach provides us with a set of pre-trained classifiers ${\left\{ {\left\{f_{\alpha}^{(g)}(\cdot) \right\}}_{g=1}^G  \right\}}_{a=1}^{M} $ that can be readily utilized on any new data instances without the need for those samples to be labeled by the original workers.
    \end{enumerate}
    \item Lets Define $t_{\alpha}^{(i,k),(g)} $ as the predicted label obtained by binarizing the predicted probabilities $p_{\alpha}^{ (i,k),(g)} $ using the threshold $\theta_{\alpha}^{(k),(g)} $ as shown in the Glossary of Symbols section.
    \item Uncertainty measures are used to quantify the level of uncertainty or confidence associated with the predictions of a model.
\end{itemize}
In this work, we need to measure the uncertainty $u_{\alpha}^{(i,k)}$ associated with the model predictions. Some common uncertainty measurement measures are as follows.
\subsubsection{Entropy}
Entropy is a widely used measure of uncertainty in classification problems. In an ensemble of classifiers, entropy serves as a quantitative measure of the uncertainty or disorder present in the probability distribution of the predicted class labels. A higher entropy value indicates a greater degree of uncertainty in the predictions, as the predictions of the individual classifiers in the ensemble are significantly different. In contrast, a lower entropy value indicates reduced uncertainty as the ensemble assigns very similar probabilities to a particular class, indicating strong agreement among the classifiers and increased confidence in their collective prediction. The formula for calculating entropy is as follows:
\begin{align}
    \Delta_{\alpha}^{(i,k)}
    \, & = \, \textcolor{gray}{H\left({\left\{p_{\alpha}^{(i,k),(g)}\right\}}_{g=1}^{G}\right)} \\
    \, & = \, -\sum_{g}{p_{\alpha}^{(i,k),(g)} \log\left(p_{\alpha}^{(i,k),(g)}\right)}
    \label{eq:crowd.Eq.5.uncertainty}
\end{align}
\subsubsection{Standard Deviation}
In regression problems, standard deviation is often used to quantify uncertainty. It measures the dispersion of predicted values around the mean. A greater standard deviation indicates greater uncertainty of the prediction. For a set of predicted values ${\{t_{\alpha}^{(i,k),(g)} \}}_{g=1}^G $ with mean value $\mu $, the standard deviation is defined as.
\begin{align}
\Delta_{\alpha}^{(i,k)}
\, & = \, \textcolor{gray}{\text{SD}\left({\left\{t_{\alpha}^{(i,k),(g)}\right\}}_{g=1}^G\right)} \\
\, & = \, \sqrt {\frac{1}{G-1}\sum_{g=1}^G {\left(t_{\alpha}^{(i,k),(g)}-\mu\right)}^2}
\label{eq:crowd.Eq.6.uncertainty.sd}
\end{align}
where $\mu=\frac{1}{G}\sum_{g=1}^{G}{t_{\alpha}^{(i,k),(g)}}$.

\subsubsection{Predictive Interval}
A predictive interval provides a range within which a future observation is likely to fall with a certain level of confidence. For example, a 95\% predictive interval indicates that there is a 95\% likelihood that the true value falls within that range. A greater uncertainty corresponds to wider intervals. In the context of multiple classifiers, the predictive intervals can be calculated by considering the quantiles of the classifier output. For a predefined confidence level $\gamma $ (e.g., 95\%), for a specific class $k $, we need to find the quantiles $Q_{L}^{k} $ and $Q_{U}^{k} $ of the probability distribution of class $k $ predicted by the $G $ classifiers. The uncertainty can be represented by the width of the predictive interval:
\begin{align}
    P\left(Q_L^{k} \leq p_{\alpha}^{(i,k),(g)} \leq Q_U^{k}\right) & = \gamma
    \\
    \Delta_{\alpha}^{(i,k)} & = Q_L^{k} - Q_U^{k}
    \label{eq:crowd.Eq.uncertainty}
\end{align}
The steps to calculate the predictive interval are as follows:
\begin{enumerate}
    \item Collect the class $k $ probabilities predicted by all $G$ classifiers for a given instance. Then sort the values in ascending order. Let us call this set $P_{\alpha}^{(i,k)}=\mathrm{sorted}\left({\left\{p_{\alpha}^{(i,k),(g)}\right\}}_{g=1}^G\right),\quad\forall \alpha,k,i $.
    \item Calculate the lower and upper quantile indices based on the chosen confidence level $\gamma $. The lower quantile index is $L=\mathrm{ceil}\left(\frac{G}{2}\left(1-\gamma\right)\right) $, and the upper quantile index is $U=\mathrm{floor}\left(\frac{G}{2} (1+\gamma)\right) $, where ceil and floor are the ceiling and floor functions, respectively.
    \item Find the values corresponding to the lower and upper quantile indices in the sorted $P_{\alpha,k}^{(i)} $. These values are the lower and upper quantiles $Q_L^{k} $ and $Q_U^{k} $.
    \item Now we have the predictive interval $P\left(Q_L^{k}<=p_{\alpha}^{(i,k),(g)}<=Q_U^{k}\right)=\gamma $, where $Q_L^{k} $ and $Q_U^{k} $ represent the bounds of the interval containing the $\alpha$ proportion of the probability mass.
\end{enumerate}
\subsubsection{Monte Carlo Dropout}
The Monte Carlo dropout~\cite{gal_Dropout_2016a} can be used to estimate uncertainty in neural networks by applying the dropout at test time. Multiple forward passes with dropout generate a distribution of predictions from which uncertainty can be derived using any of the aforementioned techniques (standard deviation, entropy, etc.).
\subsubsection{Bayesian Approaches}
Bayesian methods offer a probabilistic framework to estimate the parameters of the model and make predictions. These methods explicitly model uncertainty by considering prior beliefs about the model parameters and then updating those beliefs based on the observed data. In Bayesian modeling, the model parameters are treated as random variables and a posterior distribution is estimated using these parameters. The following are two common Bayesian approaches for measuring the uncertainty in classification problems.
\begin{itemize}
    \item \relax \textbf{Bayesian model averaging (BMA):} BMA accounts for model uncertainty by combining the predictions of various models using their posterior probabilities as weighting factors. Instead of selecting a single ``best'' model, BMA acknowledges the possibility of multiple plausible models, each with its own strengths and weaknesses~\cite{hoeting_Bayesian_1999}. The steps to implement BMA are as follows. Select a set of candidate models that represent different hypotheses regarding the data-generating process underlying the data. These models may be of various types, such as linear regression, decision trees, neural networks, or any other model suited to the specific problem at hand. Using the available data, train each candidate model. Calculate the posterior probabilities of the models. Using the posterior probabilities of each model as weights, calculate the weighted average of each model's predictions. The weighted average is the BMA prediction for the input instance and class.
    \item \relax \textbf{Bayesian neural networks (BNNs):} BNNs~\cite{mullachery_Bayesian_2018} are an extension of conventional neural networks in which the weights and biases of the network are treated as random variables. The primary distinction between BNNs and conventional neural networks is that BNNs model uncertainty directly in the weights and biases. The posterior distributions of the network weights and biases (learned during training) capture the uncertainty, which can then be utilized to generate predictive distributions for each class. This enables multiple predictions to be generated by sampling these predictive distributions, which can be used to quantify the uncertainty associated with each class.
\end{itemize}
\subsubsection{Committee-Based Methods}
The committee-based method~\cite{wang_Wisdom_2020} involves training multiple models (a committee) and aggregating their predictions. The disagreement between committee members' predictions can be used as a measure of uncertainty. Examples include bagging and boosting ensemble methods and models, such as random forests.
\begin{align}
    \Delta_{\alpha}^{(i,k)}
    \, & = \, \textcolor{gray}{\mathrm{VarCommittee}\left(P_{\alpha}^{(i,k)}\right)} \\
    \, & = \, \frac{1}{G-1} \sum_{g=1}^G {\left(p_{\alpha}^{(i,k),(g)}-\mu\right)}^{2}
    \label{eq:crowd.Eq.8.uncertainty.committee_based}
\end{align}
where $\mu= \frac{1}{G} \sum_{g=1}^G p_{\alpha}^{(i,k),(g)}$.
\subsubsection{Conformal Prediction}
Conformal prediction~\cite{angelopoulos_Gentle_2021} is a method of constructing prediction regions that maintain a predefined level of confidence. These regions can be used to quantify the uncertainty associated with the prediction of a model.

Steps  to calculate the nonconformity score:
\begin{enumerate}
    \item For each classifier $g $ and each class $k $, calculate the nonconformity score. Here, $\mathrm{score\_function}$ measures the conformity of the prediction with the true label. In the context of this study, the true label can be replaced by $\eta_{\alpha}^{(i,k)} $. A common choice for $\mathrm{score\_function}$ is the absolute difference between the predicted probability and the true label, but other options can be used depending on the specific problem and requirements. Define the nonconformity score as $ \zeta_{k}^{g} = \mathrm{score\_function} \left(p_{\alpha}^{(i,k),(g)}, y^{(i,k)}\right) $
    \item Calculate the p-value for each class $k $ as the proportion of classifiers with nonconformity scores greater than or equal to a predefined threshold $\text{T}^{(k)}: \text{p-values}(k) = \frac{ \left\vert \{g: \;\zeta^{(k),(g)} \geq \text{T}^{(k)} \} \right\vert} {G} $
    \item The p-values calculated for each class $k $ represent the uncertainty associated with that class. A higher p-value indicates a higher level of agreement among the classifiers for a given class, whereas a lower p-value suggests greater uncertainty or disagreement.
\end{enumerate}
The uncertainty measures discussed above are only some of the available options. Selecting an appropriate measure depends on factors such as the problem domain, the chosen model, and the specific requirements of a given application. For this study, we use the variance technique shown in Equation~(\ref{eq:crowd.Eq.6.uncertainty.sd}) as our uncertainty measurement due to its simplicity. However, other measures could also be employed as suitable alternatives.
\subsection{Crowd-Certain: Uncertainty-Based Weighted Soft Majority Voting}
\subsubsection{Consistency Measurement}
Define $c_{\alpha}^{(i,k)} $ as the consistency score for worker $\alpha $, class $k $ and instance $i $. We calculate this consistency score using the uncertainty score $\Delta_{\alpha}^{(i,k)} $ explained in the previous section. We use two approaches to calculate $c_{\alpha}^{(i,k)} $ from $\Delta_{\alpha}^{(i,k)} $.
\paragraph*{Method 1: Consistency Measurement without Penalization}
The first approach is to simply subtract the uncertainty from $1 $ as follows:
\begin{equation}
    c_{\alpha}^{(i,k)}=1-\Delta_{\alpha}^{(i,k)}\;\;,\;\forall i,\alpha,k
    \label{eq:crowd.Eq.9.consistency}%
\end{equation}
\paragraph*{Method 2: Consistency Measurement with Penalization}
In a second approach (shown in Equation~(\ref{eq:crowd.eq.10.consistency-penalized})), we penalize workers for instances in which their predicted label $\eta_{\alpha}^{(i,k)} $ (explained in the Glossary of Symbols section) does not match the MV of all worker labels ${{\underset \alpha{\mathrm{MV}}}{\left(z_{\alpha,\;k}^{(i,k)}\right)}} $. As previously discussed, instead of directly working with the worker's labels $z_{\alpha}^{(i,k)} $, we use the predicted labels obtained from the ensemble of classifiers $\eta_{\alpha}^{(i,k)} $. This methodology does not require repeating the crowd-labeling process for new data samples. In particular, we are likely not to have access to the same crowd of workers employed in the training dataset.
\begin{equation}
    c_{\alpha}^{(i,k)} =
    \begin{cases}
        1 - \Delta_{\alpha}^{(i,k)} & \text{if } \eta_{\alpha}^{(i,k)} = \operatorname{MV}_{\alpha}(\eta_{\alpha}^{(i,k)}) \\
        0 & \text{otherwise}
    \end{cases}
    \label{eq:crowd.eq.10.consistency-penalized}
\end{equation}
\subsubsection{Reliability Measurement}
For each worker, for each class, and for each instance, there is a consistency score, $c_{\alpha}^{(i,k)}$. By averaging these scores across all instances, we can define a reliability score for each worker and for each class:
\begin{equation}
    \psi_{\alpha}^{(k)}=\frac{1}{N}\sum_{i=1}^N c_{\alpha}^{(i,k)}\label{eq:crowd.reliability_measurement}
\end{equation}
If desired, one may also calculate an overall reliability score for each worker by averaging across all classes:
\begin{equation}
    \psi_{\alpha}=\frac{1}{K}\sum_{k=1}^K \psi_{\alpha}^{(k)}
\end{equation}
\subsubsection{Weight Measurement}
Furthermore, we calculate the workers' weights $\omega_\alpha^{(k)}$ for each class k by normalizing the reliability values as follows:
\begin{equation}
    \omega_{\alpha}^{(k)}=\frac{\psi_{\alpha}^{(k)}}{\sum_{\alpha=1}^{M} \psi_{\alpha}^{(k)}}
    \label{eq:crowd.eq.11.weights}
\end{equation}
\subsubsection{Aggregated Label Calculation}
Finally, the aggregated label $\nu^{(i,k)}$ for each instance $i $ and class $k $ is the weighted average of the predicted labels $\eta_{\alpha}^{(i,k)} $ for each worker $\alpha $:
\begin{equation}
    \nu^{(i,k)} =
    \begin{cases}
        1 & \text{if } \left(\sum_{\alpha=1}^{M} \omega_{\alpha}^{(k)}\, \eta_{\alpha}^{(i,k)}\right) > 0.5 \\
        0 & \text{otherwise}
    \end{cases}
    \quad \forall i, k
    \label{eq:crowd.aggregated_label}
\end{equation}

\subsubsection{Confidence Score Calculation}
In the previous section, we showed how to calculate the aggregated label $\nu^{(i,k)}$ (shown in Equation~(\ref{eq:crowd.aggregated_label})). In this section, we use the aggregated label $\nu^{(i,k)}$ to calculate a confidence score (in the range 0 to 1) for each instance $i $ and class $k $. We calculate two confidence scores ($F_{\Omega}^{(i,k)} $ and $F_{\beta}^{(i,k)} $), based on how many different workers agree on the aggregated label $\nu^{(i,k)}$. The confidence score show the level of confidence we should place on the aggregated labels. To calculate this confidence score, we modify the two techniques used by Sheng~\cite{sheng_Majority_2019} and Tao~\cite{tao_Label_2020} to incorporate our calculated weight $\omega_{\alpha}^{(k)} $ shown in Equation~(\ref{eq:crowd.eq.11.weights})  for each worker $\alpha $.

\paragraph*{Method 1: Confidence Score Calculation using Weighted Sum}
In a standard voting system, for every instance $i$ and class $k$, each contributor in the group—whether a worker in a crowd or a model in an ensemble learning context—provides a class label. The label receiving the most votes, meaning it is predicted by the majority of contributors, is selected as the final prediction. This approach is often referred to as majority voting or hard voting.
This method could be improved by taking into account not only the number of votes each label receives, but also the confidence associated with each vote. This introduces the notion of a ``weighted sum of all votes for a particular class''. As part of this study, we propose techniques that assign a weight to each contributor in the group. This calculation is based on their voting consistency and the degree to which they concur with their peers.
To compute the weighted sum of all votes for each class, we can combine the calculated weights with the corresponding labels, whether provided or predicted. This calculation gives greater confidence to votes with greater certainty. This refined approach prioritizes certainty, thereby enhancing the ensemble's overall effectiveness.
The confidence score $F_{\Omega}^{(i,k)}$ is formulated as follow.
\begin{equation}
    F^{(i,k)} = F_{\Omega}^{(i,k)} = {\sum\nolimits_{\alpha=1}^{M}{\omega_{\alpha}^{(k)} \delta\left(\eta_{\alpha}^{(i,k)} \;,\;\; \nu^{(i,k)} \right)}}
    \label{eq:crowd.Eq.13.confidence-score.Freq}
\end{equation}
where $\delta $ is the Kronecker delta function.

\paragraph*{Method 2: Confidence Score Calculation using CDF of Beta Distribution Function}
The binomial distribution survival function, also referred to as the complementary cumulative distribution function (CCDF), provides the probability of observing a result as extreme or more extreme than a given value. It provides the probability that a random variable drawn from the binomial distribution is greater than or equal to a given value. It can be applied in the following ways when calculating confidence scores:
\begin{enumerate}
    \item \textbf{Hypothesis testing:}
    Suppose we want to test a hypothesis concerning a parameter of a population, using a set of observed samples. Given the null hypothesis, the binomial survival function can be used to calculate the probability of observing a result as extreme or more extreme than the one observed. This probability is the p-value, and if it is very small, the null hypothesis may be rejected. In this context, the confidence score could be interpreted as (1 $-$ p-value), a measure of the certainty with which the null hypothesis can be rejected.
    \item \textbf{Binary classification:} Consider a binary classification problem in which an algorithm sorts objects into two groups. Given the observed results, the binomial survival function can be used to calculate the probability of misclassification. The confidence score in this instance could be interpreted as (1 $-$ probability of misclassification).
\end{enumerate}
The confidence score serves as a quantitative measure of the reliability of the classification of instance `i' into class `k'. The CDF of the beta distribution function can be defined as follows~\cite{BetaDistribution2023a}:
\begin{equation}
    F(x; l, u) = \frac{B(x; l, u)}{B(l, u)} = I_{x} (l, u)
    \label{eq:crowd.Eq.beta_distribution_function}
\end{equation}
where $B(x; l, u)$ is the incomplete beta function, and $I_{x}(l,u)$ is the regularized incomplete beta function. $l$ and $u$ terms are typically shown as $\alpha$ and $\beta$, but we use different notations in this paper to avoid confusion with the annotator index $\alpha$ and confidence score measurement $\beta$.

The CDF of the beta distribution at the decision threshold of $x=0.5 $ (denoted as $I_{0.5}$) is used to calculate a confidence score $F_{\beta}^{(i,k)}$. To calculate $I_{0.5}$, we first need to calculate two shape parameters $l^{(i,k)}$ and $u^{(i,k)}$
\paragraph{\textbf{Shape Parameters:}}
Two shape parameters $l^{(i,k)}$ and $u^{(i,k)}$ of the Beta distribution are calculated as follows~\cite{tao_Label_2020}:
\begin{equation}
    \begin{aligned}
        l^{(i,k)} \, & = \, 1 + \sum_{\alpha=1}^{M} \omega_{\alpha}^{(k)} \; \delta\left(\eta_{\alpha}^{(i,k)}, \nu^{(i,k)}\right) \\
        u^{(i,k)} \, & = \, 1 + \sum_{\alpha=1}^{M} \omega_{\alpha}^{(k)} \; \delta\left(\eta_{\alpha}^{(i,k)}, 1 - \nu^{(i,k)}\right)
    \end{aligned}
    \label{eq:crowd.Eq.14.beta_l_u}
\end{equation}
A Major difference between our calculations shown in Equation~(\ref{eq:crowd.Eq.14.beta_l_u}) and Tao~\cite{tao_Label_2020}, is that, Tao calculates a weighted sum of the annotators' labels ($z_{\alpha}^{(i,k)}$) that have voted on the positive class ($\delta (z_\alpha^{(i,k)},+)$) or negative class ($\delta (z_\alpha^{(i,k)},-)$). However, we take the predicted labels obtained from the trained classifiers belonging to each annotator ($\eta_{\alpha}^{(i,k)}$) instead of annotators' labels ($z_\alpha^{(i,k)}$). Also instead of calculating the weighted sum of all positive and negative labels, we calculate the weighted sum of all annotators' predicted labels that are the same as the calculated aggregated label (Equation~(\ref{eq:crowd.aggregated_label})), ($\delta (\eta_{\alpha}^{(i,k)}, \nu^{(i,k)})$) and differs from it ($\delta (\eta_{\alpha}^{(i,k)}, 1-\nu^{(i,k)})$) respectively.
Here, the shape parameters are effectively a weighted sum (with weights $\omega_{\alpha}^{(k)}$) of all the correct and incorrect aggregated labels, modulated by a Dirac delta function $\delta$, which acts to selectively include terms where the condition inside the delta function is satisfied.
\paragraph{\textbf{Confidence Score:}}
The confidence score is subsequently calculated utilizing the previously determined shape parameters, as follows:
\begin{align}
    F^{(i,k)} = F_{\beta}^{(i,k)}
    \, & = \, \textcolor{gray}{I_{0.5}\left(l^{(i,k)},u^{(i,k)}\right)} \\
    \, & = \, \sum_{t=[l^{(i,k)}]}^{T-1}\frac{(T-1)!}{t!(T-1-t)!}0.5^{T-1}
    \label{eq:crowd.Eq.15.confidence-score.Beta}
\end{align}
where $T = \left [ l^{(i,k)} + u^{(i,k)} \right ] $ and $\left [ \cdot \right ] $ denotes rounding to the nearest integer.
\subsection{Metrics}
\begin{itemize}
    \item \textbf{Accuracy:} The accuracy of the model is the proportion of true results (both true positive and true negatives) among the total number of cases examined. Mathematically, accuracy can be represented as
    \begin{equation}
        \text{Accuracy}= \frac{1}{N} \sum_{i=1}^{N} \sum_{k=1}^{K} \delta\left(\nu^{(i,k)}, y^{(i,k)}\right)
    \end{equation}
    where $\delta$ is the Kronecker delta function, $N$ is the total number of instances, and $K$ is the number of classes, and $y^{(i,k)}$ and $\nu^{(i,k)}$ are the ground truth and aggregated label, respectively, for class $k$ and instance $i$. Accuracy is most effective for balanced classes, and its interpretation can be skewed in the presence of significant class imbalance.
    \item \textbf{F1 Score:} The F1 score is the harmonic mean of precision and recall and can be used for assessing the quality of aggregated labels, especially in the presence of imbalanced classes. F1 score provides a balanced measure of precision and recall, ranging from 0 to 1, where 1 represents the best possible F1 score. It is computed as
    \begin{equation}
        \text{F1} = 2 \cdot \frac{\text{Precision} \cdot \text{Recall}}{\text{Precision} + \text{Recall}}
    \end{equation}
    where $\text{Precision} = \frac{\text{TP}}{\text{TP+FP}}$ and $\text{Recall} = \frac{\text{TP}}{\text{TP+FN}}$, and TP, FP and FN are the numbers of true positives, false positives and false negatives, respectively.
    \item \textbf{Area Under the Curve for the Receiver Operating Characteristic (AUC-ROC):} AUC-ROC measures the trade-off between the true positive rate (sensitivity) and false positive rate (1-specificity). Higher AUC-ROC values indicate better classification performance.
    \item \textbf{Brier Score:} Brier score provides a measure of the accuracy of the probabilistic (or confidence score) predictions. It is calculated as the mean squared error between the estimated confidence score $F^{(i,k)}$ and the ground truth label $y^{(i,k)}$. thereby rewarding more calibrated predictions. Calibration here is defined as the alignment between measured confidence scores and their corresponding observed frequencies. Brier Score can be calculated as follows:
    \begin{equation}
        \text{Brier Score} = \frac{1}{N} \sum_{i=1}^{N} \sum_{k=1}^{K} {\left(F^{(i,k)} - y^{(i,k)}\right)}^2
    \end{equation}
where $F^{(i,k)}$ can be any probabilistic measure such as $F_{\beta}^{(i,k)}$ or $F_{\Omega}^{(i,k)}$.
    \item \textbf{Expected Calibration Error (ECE):}  The ECE quantifies the calibration of  confidence scores produced by a model. It is computed as a weighted average of the absolute differences between the actual accuracies and the predicted confidences within each bin when predictions are grouped into distinct bins based on their predicted confidence. A lower ECE signifies a model whose predicted probabilities closely match the observed frequencies across all bins. ECE can be formulated as follows:
    \begin{equation}
        \text{ECE} = \sum_{b=1}^{B} \frac{ \vert B_b \vert }{N} \left\vert {\text{Accuracy} ( B_{b} ) - \text{Confidence-Score}(B_b)} \right\vert
    \end{equation}
    where $B$ is the number of bins, $B_b$ is the set of instances in bin $b$, $N$ is the total number of instances, $\text{Accuracy} (B_b)$ is the accuracy of bin $b$, and $\text{Confidence-Score} (B_b)$ is the average confidence of bin $b$.
\end{itemize}
\section{Results}\label{sec:crowd.results}
To evaluate our proposed technique, we conducted a series of experiments comparing the proposed technique with several existing techniques such as MV, Tao~\cite{tao_Label_2020}, and Sheng~\cite{sheng_Majority_2019}, as well as with other crowdsourcing methodologies reported in the crowd-kit package~\cite{ustalov_learning_2021} including Gold Majority Voting, MMSR~\cite{ma_Adversarial_2020}, Wawa, Zero-Based Skill, GLAD~\cite{whitehill_Whose_2009}, and Dawid Skene~\cite{dawid_Maximum_1979}.
\subsection{Datasets}\label{subsec:results.datasets}
We report the performance of our proposed techniques on various datasets. These datasets cover a wide range of domains and have varying characteristics in terms of the number of features, samples, and class distributions. Table~\ref{tab:crowd.Table.1.Datasets} provides an overview of the datasets used. All datasets are obtained from the University of California, Irvine (UCI) repository~\cite{duan_UCI_2017}.
\begin{table}[htbp]
\centering
\caption[Description of the Datasets Used]{Description of the datasets used.}\label{tab:crowd.Table.1.Datasets}
\begin{tabulary}{\linewidth}{LCCCC}
    \toprule
    \textbf{Dataset} & \textbf{\#Features} & \textbf{\#Samples} & \textbf{\#Positives} & \textbf{\#Negatives} \\
    kr-vs-kp    & 36 & 3196 & 1669 & 1527 \\
    mushroom    & 22 & 8124 & 4208 & 3916 \\
    iris        & 4  & 100  & 50   & 50   \\
    spambase    & 58 & 4601 & 1813 & 2788 \\
    tic-tac-toe & 10 & 958  & 332  & 626  \\
    sick        & 30 & 3772 & 231  & 3541 \\
    waveform    & 41 & 5000 & 1692 & 3308 \\
    car         & 6  & 1728 & 518  & 1210 \\
    vote        & 16 & 435  & 267  & 168  \\
    ionosphere  & 34 & 351  & 126  & 225  \\
    \bottomrule
\end{tabulary}
\end{table}
\begin{itemize}
    \item The kr-vs-kp dataset represents the King Rook-King Pawn on a7 in chess. The positive class indicates a victory for white (1,669 instances, or 52\%), while the negative class indicates a defeat for white (1,527 instances, 48\%).
    \item The mushroom dataset is based on the Audubon Society Field Guide for North American Mushrooms (1981) and includes 21 attributes related to mushroom characteristics such as cap shape, surface, odor, and ring type.
    \item The Iris plants dataset comprises three classes, each with 50 instances, representing different iris plant species. The dataset contains four numerical attributes in centimeters: sepal length, sepal width, petal length, and petal width.
    \item The Spambase dataset consists of 57 attributes, each representing the frequency of a term appearing in an email, such as the ``address''.
    \item The tic-tac-toe endgame dataset encodes all possible board configurations for the game, with ``x'' playing first. It contains attributes (X, O, and blank) corresponding to each of the nine tic-tac-toe squares.
    \item The Sick dataset includes thyroid disease records from the Garvan Institute and J. Ross Quinlan of the New South Wales Institute in Sydney, Australia. 3,772 instances with 30 attributes (seven continuous and 23 discrete) and 5.4\% missing data. Attributes include age, pregnancy, TSH, T3, TT4, etc.
    \item The waveform dataset generator comprises 41 attributes and three wave types, with each class consisting of two ``base'' waves.
    \item The Car Evaluation Dataset rates cars on price, buying, maintenance, comfort, doors, capacity, luggage, boot size, and safety using a simple hierarchical decision model. The dataset consists of 1,728 instances categorized as unacceptable, acceptable, good, and very good.
    \item The 1984 US Congressional Voting Records dataset shows how members voted on 16 CQA-identified critical votes. Votes are divided into nine categories, simplified to yea, nay, or unknown disposition. The dataset has two classes: Democrats (267) and Republicans (168).
    \item The Johns Hopkins Ionosphere dataset contains data collected near Goose Bay, Labrador, using a phased array of 16 high-frequency antennas. ``Good'' radar returns show ionosphere structure, while ``bad'' returns are ionosphere-free. The dataset includes 351 instances with 34 attributes categorized as good or bad.
\end{itemize}
All datasets were transformed into a two-class binary problem for comparison with existing benchmarks. For instance, only the first and second classes were used in the ``waveform'' dataset, and the first two classes were utilized in the ``Iris'' dataset.
We generated multiple fictitious label sets for each dataset to simulate the crowdsourcing concept of collecting several crowd labels for each instance. We selected random samples in the datasets using a uniform distribution and altered their corresponding true labels to incorrect ones, while maintaining the original distribution of the ground-truth labels. The probability of each instance containing the correct true label was determined using a uniform distribution, allowing us to create synthetic label sets for each worker that preserved the underlying structure and difficulty of the original classification problem. By creating datasets with various levels of accuracy, we can evaluate the performance of the proposed method under different conditions of worker expertise and reliability. This allows us to assess the ability of our method to handle diverse real-world crowdsourcing scenarios and gain insight into its general applicability and effectiveness in improving overall classification accuracy.
\subsection{Benchmarks}
Tao~\cite{tao_Label_2020} and Sheng~\cite{sheng_Majority_2019} techniques were implemented in Python to evaluate their performance. Furthermore, the crowd-kit package (A General-Purpose Crowdsourcing Computational Quality Control Toolkit for Python)~\cite{ustalov_learning_2021} was used to implement the remaining benchmark techniques, including Gold Majority Voting, MMSR~\cite{ma_Adversarial_2020}, WAWA~\cite{crowdkit_webpage_documentation}, Zero-Based Skill, GLAD~\cite{whitehill_Whose_2009}, and Dawid Skene~\cite{dawid_Maximum_1979}.
\begin{itemize}
    \item \textbf{Worker Agreement with Aggregate (WAWA)~\cite{crowdkit_webpage_documentation}:} Wawa, also referred to as `inter-rater agreement', is a metric used in crowdsourcing jobs that do not employ test questions~\cite{appen_wawa_2023}. The WAWA algorithm consists of three steps: it calculates the majority vote label, estimates workers' skills as a fraction, and calculates the agreement between workers and the majority vote~\cite{crowdkit_webpage_documentation}.
    \item \textbf{Zero-Based-Skill (ZBS)~\cite{crowdkit_webpage_documentation}:} employs a weighted majority vote (WMV). After processing a collection of instances, it re-evaluates the abilities of the workers based on the accuracy of their responses. This process is repeated until the labels no longer change or the maximum number of iterations is reached.
    \item \textbf{Karger-Oh-Shah (KOS)~\cite{crowdkit_webpage_documentation}:} Iterative algorithm that calculates the log-likelihood of the task being positive while modeling the reliabilities of the workers. Let $A_{(i,\alpha)} $ be a matrix of answers of worker $\alpha $ on task $i $. $A_{(i,\alpha)} = 0 $ if worker $\alpha $ didn't answer the task $i $ otherwise $\vert A_{(i,\alpha)} \vert = 1 $. The algorithm operates on real-valued task messages $x_{i \rightarrow \alpha} $  and worker messages $y^{\alpha \rightarrow i} $. A task message $x_{i \rightarrow \alpha} $ represents the log-likelihood of task $i $ being a positive task, and a worker message $y_{\alpha \rightarrow i} $ represents how reliable worker $\alpha $ is. On iteration $k$ the values are updated as follows~\cite{crowdkit_webpage_documentation}:
    \begin{equation}
        x_{i \rightarrow \alpha}^{(k)} = \sum_{\alpha' \in \partial i \backslash \alpha} A_{(i, \alpha')} y_{\alpha' \rightarrow i}^{(k-1)} \\
        y_{\alpha \rightarrow i}^{(k)} = \sum_{i' \in \partial \alpha \backslash i} A_{(i',\alpha)} x_{i' \rightarrow \alpha}^{(k-1)}
    \end{equation}
    \item \textbf{Multi-Annotator Competence Estimation (MACE)~\cite{hovy_MACE_2013,crowdkit_webpage_documentation}:}
    Probabilistic model that associates each worker with a probability distribution over the labels. For each task, a worker might be in a spamming or not spamming state. If the worker is not spamming, they yield a correct label. If the worker is spamming, they answer according to their probability distribution. Let's assume that the correct label $y^{(i)}$ comes from a discrete uniform distribution. When a worker annotates the task, they are in the spamming state with probability $\operatorname{Bernoulli}(1 - \theta_{\alpha})$. So, if their state $s_{\alpha} = 0$, their response is $z^{(i)}_{\alpha} = y^{(i)}$. Otherwise, their response $z^{(i,\alpha)}$ is drawn from a multinomial distribution with parameters $\xi_{\alpha}$.
    \item \textbf{Matrix Mean-Subsequence-Reduced Algorithm (MMSR)~\cite{ma_Adversarial_2020,crowdkit_webpage_documentation}:} The MMSR assumes that workers have different levels of expertise and are associated with a vector of ``skills'' $\boldsymbol{s}$ which has entries $s_{\alpha}$ showing the probability that the worker $\alpha$ answers correctly to the given task. Having that, we can show that%
    \begin{equation}
        \mathbb{E}\left[\frac{K}{K-1}\widetilde{C}-\frac{1}{K-1}\boldsymbol{1}\boldsymbol{1}^T\right]
        = \boldsymbol{s}\boldsymbol{s}^T,
    \end{equation}
    where $K$ is the total number of classes, $\widetilde{C}$ is a covariance matrix between workers, and $\boldsymbol{1}\boldsymbol{1}^T$ is the all-ones matrix, which has the same size as $\widetilde{C}$.
    So, the problem of recovering the skills vector $\boldsymbol{s}$ becomes equivalent to the rank-one matrix completion problem. The MMSR algorithm is an iterative algorithm for rank-one matrix completion, so its result is an estimator of the vector $\boldsymbol{s}$. Then, the aggregation is the weighted majority vote with weights equal to $\log \frac{(K-1)s_{\alpha}}{1-s_{\alpha}}$.
    \item \textbf{Generative model of Labels, Abilities, and Difficulties (GLAD)~\cite{whitehill_Whose_2009,crowdkit_webpage_documentation}:} A probabilistic model that parameterizes workers' abilities and tasks' difficulties. Let's consider a case of $K$ class classification. Let $p$ be a vector of prior class probabilities, $\omega_{\alpha} \in (-\infty, +\infty)$ be a worker's ability parameter, $\beta^{(k)} \in (0, +\infty)$ be an inverse task's difficulty, $y^{(k)}$ be a latent variable representing the true task's label, and $z_{\alpha}^{(k)}$ be a worker's response that we observe. The relationships between these variables and parameters according to GLAD are represented by the following latent label model.
    The prior probability of $y^{(k)}$ being equal to $c$ is $\Pr(y^{(k)} = c) = p[c]$,
    the probability distribution of the worker's responses conditioned by the true label value $c$ follows the single coin Dawid-Skene model, where the true label probability is a sigmoid function of the product of the worker's ability and the inverse task's difficulty:
    \begin{equation}
        \Pr \left( z^{(k)}_\alpha = j \, \vert \, y^{(k)} = c \right) = \begin{cases}f(\alpha, k), & j = c \\ \frac{1 - f(\alpha, k)}{K-1}, & j \neq c \end{cases}
    \end{equation}
    where $f(\alpha, k) = \frac{1}{1 + e^{-\omega_{\alpha} \beta^{(k)} }}$.
    Parameters $p$, $\omega$, $\beta$ and latent variables $y$ are optimized through the expectation-minimization algorithm.
    \item \textbf{Dawid-Skene~\cite{dawid_Maximum_1979,crowdkit_webpage_documentation}:}
    Probabilistic model that parameterizes workers' level of expertise through confusion matrices.  Let $e^{\alpha}$ be a worker's confusion (error) matrix of size $K \times K$ in case of $K$ class classification, $p$ be a vector of prior class probabilities, $y^{(i)}$ be a true task's label, and $z_{\alpha}^{(i)}$ be a worker's answer for the task $i$. The relationships between these parameters are represented by the following latent label model.
    Here, the prior true label probability is $\Pr(y^{(i)} = c) = p[c]$ and the distribution of the worker's responses given the true label $c$ is represented by the
    corresponding column of the error matrix: $\Pr(z_{\alpha}^{(i)} = k \vert y^{(i)} = c) = e^{\alpha}[k, c]$.
    Parameters $p$ and $e^{\alpha}$ and latent variables $z$ are optimized through the expectation-maximization algorithm.
\end{itemize}
\subsection{Weight Measurement Evaluation}
\begin{figure}[htbp]
    \centering
\includegraphics[width=\textwidth]{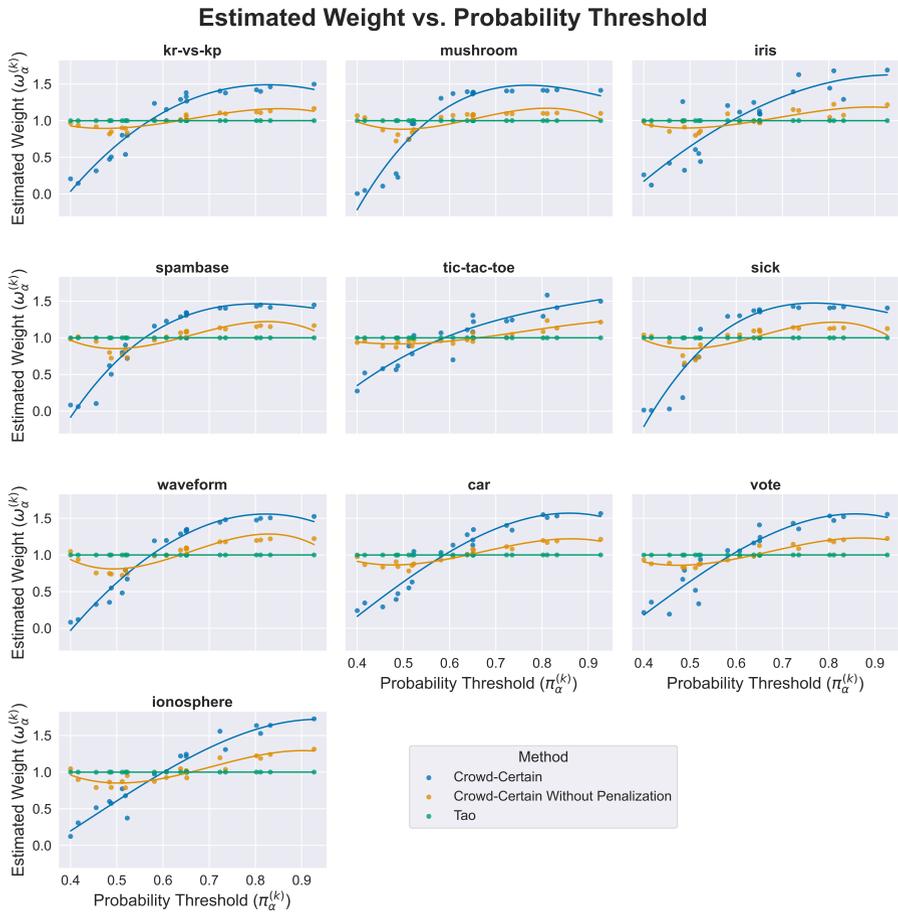}
    \caption[Comparative Analysis of Weight Computation Techniques Across Ten Datasets]{A comprehensive comparison of weight computation techniques across ten distinctive datasets. Each subplot corresponds to a specific dataset, visually illustrating the relationship between the randomly assigned worker's probability threshold ($\pi_\alpha^{(k)}$) (represented on the horizontal axis) and the resulting computed weights ($\omega_{\alpha}^{(k)}$) (shown on the vertical axis). This relationship is analyzed for the Crowd-Certain method in two scenarios --- with and without penalization --- and is also compared with the Tao method~\cite{tao_Label_2020}. Individual data points represent real measured weights, while the overlaid curve delineates the corresponding regression line.}\label{fig:crowd.weight_strength_relation}
\end{figure}

Following the generation of multi-label sets, the aggregate labels were determined using the proposed Crowd-Certain as well as various established methods. We examined two strategies for classifier selection, as detailed in Section~\ref{subsec:crowd.uncertainty}. Because there was no substantial variation in the final outcomes observed, the second strategy was adopted for its utilization of the random forest classification technique. This choice not only conserved processing time but also decreased the need for numerous Python package dependencies. For each worker $\alpha $, we trained ten distinct random forests, each comprising four trees with a maximum depth of four, under various random states, as outlined in Section~\ref{sec:crowd.method}.
Figure~\ref{fig:crowd.weight_strength_relation} depicts the relationship between the randomly assigned workers' probability threshold ($\pi_\alpha^{(k)}$) and their corresponding estimated weights ($\omega_\alpha^{(k)}$). In Tao's method scenario, the figure presents the average weights over all instances. Notably, as the reliability (probability threshold) of a worker exceeds a particular threshold, the weight computed by Tao's method reaches a saturation point, while the proposed technique exhibits a considerably stronger correlation. The individual data points symbolize the actual calculated weights, and the curve illustrates the regression line.
\subsection{Label Aggregation Evaluation}
\begin{figure}[htbp]
\centering
\includegraphics[width=\textwidth]{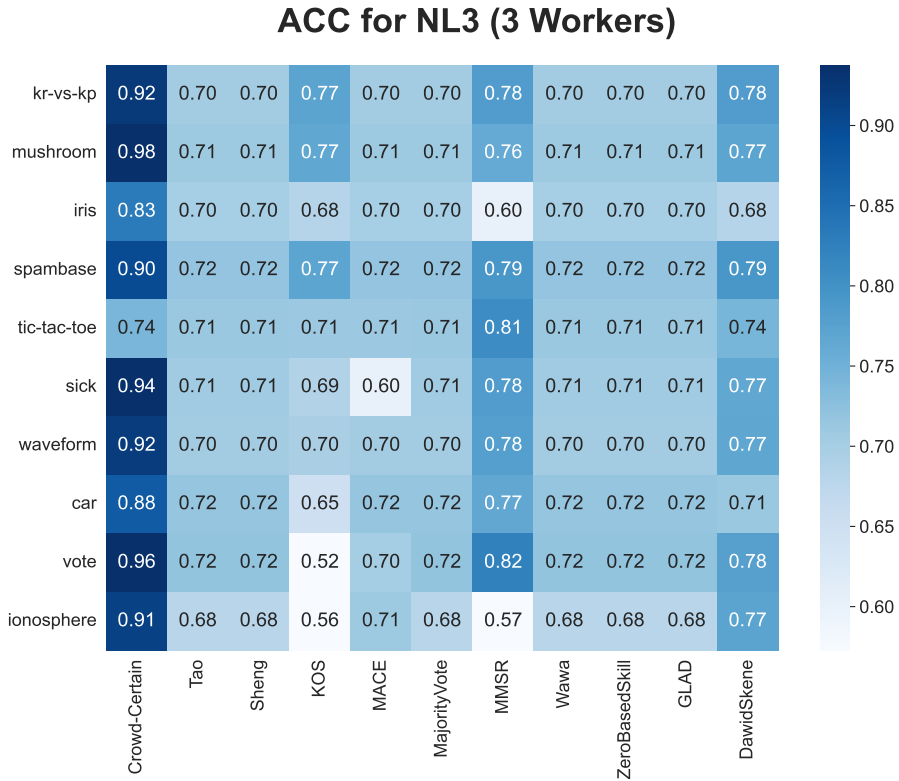}
\caption[Comparison of Mean Accuracy for Crowd-Certain and Other Aggregation Methods with Three Crowd Workers.]{Comparative analysis of the mean accuracy between the proposed Crowd-Certain method and ten pre-existing label aggregation techniques, under conditions featuring three crowd workers. The depicted mean accuracy score is derived from an averaging process across three separate trials, each initiated with a distinct random seed, thus ensuring a fair and balanced comparison. Darker blue means higher mean accuracy.}\label{fig:crowd.accuracy_per_worker}
\end{figure}

\begin{figure}[htbp]
    \centering
\includegraphics[width=\textwidth]{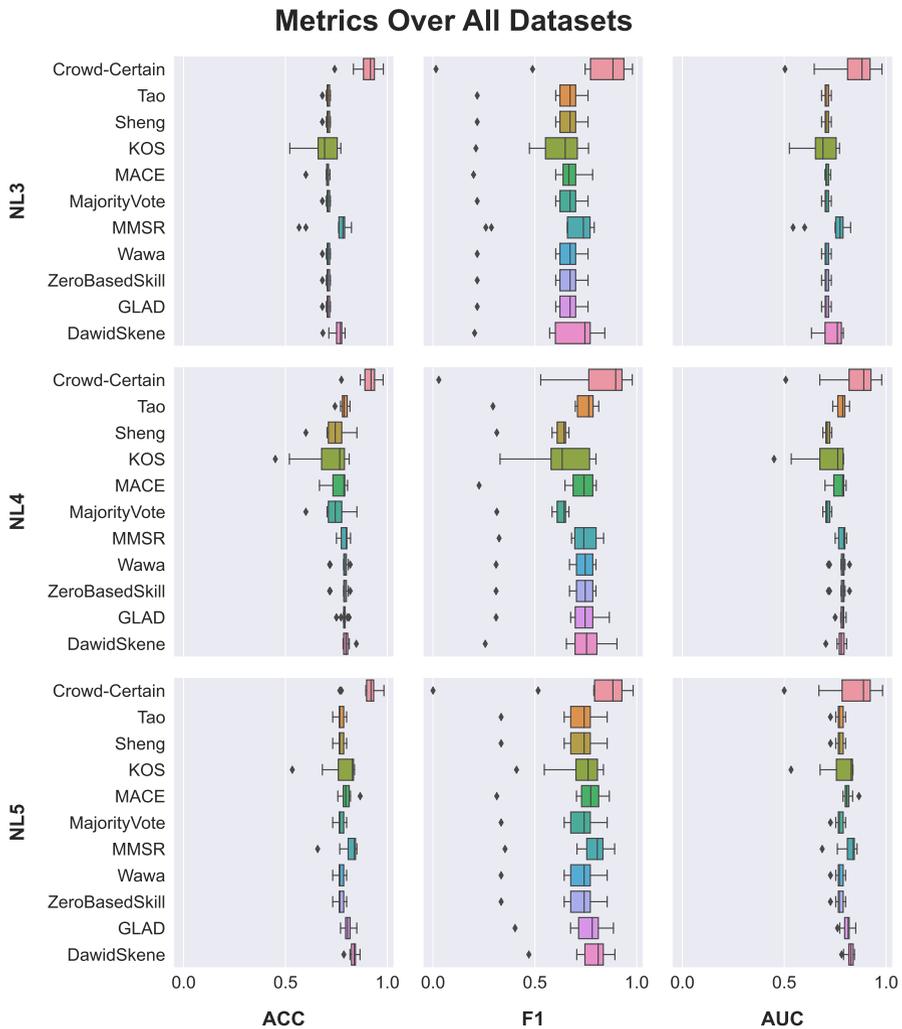}%
    \caption[Comparative Study of Accuracy, F1, and AUC Metrics for Label Aggregation Techniques Across Varied Crowd Sizes]  {This $3 \times 3$ structured figure provides a comprehensive comparison of Accuracy (first column), F1 (second column), and AUC scores (third column) across multiple label aggregation techniques (each shown with a unique color), including the proposed Crowd-Certain method and nine pre-existing techniques. Each row illustrates the results for a different number of crowd workers: the top row for three workers, the middle row for four workers, and the bottom row for five workers. Each subfigure presents ten boxplots, where each boxplot represents an aggregation technique. The metrics for each boxplot are computed from ten average scores, each corresponding to a distinct dataset. The average scores are derived from three independent trials, each with a different random seed. The aggregation of labels used in each experiment to calculate these metrics was obtained using Equation~(\ref{eq:crowd.aggregated_label}) for Crowd-Certain and Equation~(\ref{eq:crowd.aggregated_label_benchmarks}) for pre-existing techniques.}\label{fig:crowd.Fig.accuracy_all_datasets_workers}
\end{figure}

The Figure~\ref{fig:crowd.accuracy_per_worker} portrays the accuracy comparison of our label aggregation technique, termed Crowd-Certain, against ten existing methods, evaluated over ten distinct datasets. Each dataset was labeled by three different workers, with labels generated based on a uniform distribution and specific probability thresholds $\Pi_\alpha$ as explained in Section~\ref{subsec:methods.generating_fictitious_labelset}.
For a comprehensive evaluation, all experiments were repeated three times using different random seed numbers to account for randomness. The accuracy scores presented in the figure represent the average of these three runs and illustrate the degree of concordance between the aggregated label $\nu^{(i,k)}$ from each technique and the actual ground truth $y^{(i,k)}$.
It is important to note that, in the execution of our proposed technique, Crowd-Certain, the aggregated labels were derived through the application of the predicted probabilities, denoted as $\eta_{\alpha}^{(i,k)}$. This approach is significant as it enables the reuse of trained classifiers on future sample data, eliminating the need for recurrent simulation processes --- a substantial advantage in terms of computational efficiency. Conversely, the methodologies of existing techniques necessitated the use of actual crowd labels $z_\alpha^{(i,k)}$ to determine the aggregated labels. For example, in the case of Tao~\cite{tao_Label_2020} the aggregated labels were obtained using the following equation:
\begin{equation}
    \nu^{(i,k)} =
    \begin{cases}
        1 & \text{if } \left(\sum_{\alpha=1}^{M} \omega_{\alpha}^{(k)}\, z_{\alpha}^{(i,k)}\right) > 0.5 \\
        0 & \text{otherwise}
    \end{cases}
    \quad \forall i, k
    \label{eq:crowd.aggregated_label_benchmarks}
\end{equation}
These methods inherently involve re-running simulations for every new dataset, which could be computationally expensive and time-consuming. The Crowd-Certain method outperforms the existing methods, yielding higher average accuracy rates across 9 of the 10 evaluated datasets, achieving a smaller accuracy compared to only one of the 10 benchmarks (MMSR) on tic-tac-toe dataset. For example, in the `kr-vs-kp' dataset, our proposed Crowd-Certain method achieved an average accuracy of approximately 0.923, significantly exceeding the highest-performing existing method that reached an accuracy of about 0.784. This trend holds true across other datasets as well, such as `mushroom', `spambase', and `waveform', where the Crowd-Certain method achieves superior average accuracies of around 0.98, 0.90, and 0.92, respectively.
We further extended our experiment to explore the effects of varying the number of workers, ranging from 3 up to 7. The results shown in Figure~\ref{fig:crowd.Fig.accuracy_all_datasets_workers} are presented as a series of box plots, each illustrating the distribution of accuracy (1st column), F1 (2nd column), and AUC (3rd column) scores across the 10 datasets for a given number of workers. These plots provide a visual summary of our technique's performance across various settings, including the median, quartiles, and potential outliers in the distribution of accuracies. Notably, our proposed Crowd-Certain technique shows improvements over the 10 benchmark methods for different number of workers. This enhancement is evident irrespective of the number of workers involved.
\subsection{Confidence Score Evaluation}
\begin{figure}[htbp]
    \centering
    \includegraphics[width=\textwidth]{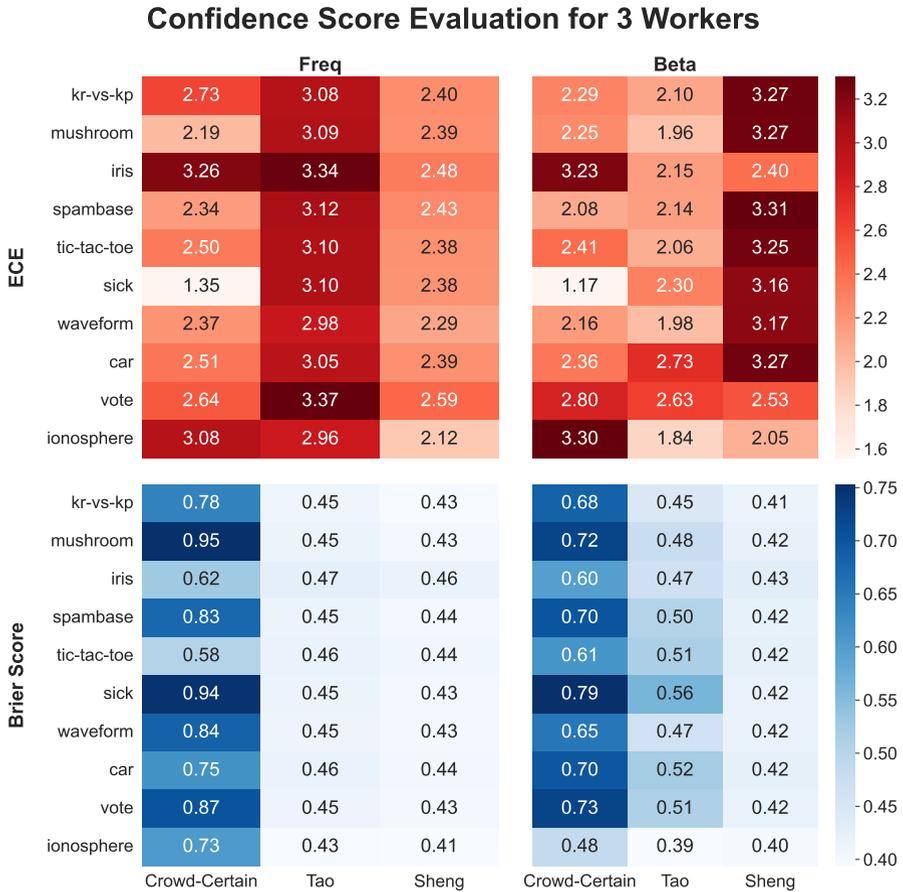}
    \caption[ECE and Brier Score Comparison across Techniques and Confidence Score Measurement Strategies for 3 Workers]{Comparative heatmap of the ECE and Brier Score across two confidence score measurement strategies: Beta ($F_{\beta}$) and Freq ($F_{\Omega}$). The comparison involves three different label aggregation techniques: Crowd-Certain, Tao, and Sheng, and spans ten distinct datasets for three crowd workers (NL3). The chosen metrics provide insight into the calibration of the predictions across different configurations}\label{fig:crowd.Fig.confidence_scores_all_datasets_3_workers}
\end{figure}

The Figure~\ref{fig:crowd.Fig.confidence_scores_all_datasets_3_workers} presents the evaluation of the two confidence score measurement techniques, namely Freq and Beta, using two performance metrics: Expected Calibration Error (ECE) and Brier Score. The evaluations were conducted across a variety of datasets and using three techniques: Crowd-Certain, Tao, and Sheng, when using three workers.
Figure~\ref{fig:crowd.Fig.confidence_scores_all_datasets_3_workers} depicts the performance of three different strategies: Crowd-Certain, Tao, and Sheng, compared across two metrics: ECE and Brier Score. These results are obtained using two different confidence score calculation techniques, Freq and Beta, applied over ten different datasets when three workers. The ECE offers a measure of how calibrated the  confidence scores ($F_{\Omega}^{(i,k)}$ and $F_{\beta}^{(i,k)}$) are across different techniques and strategies. Calibration here is defined as the alignment between measured confidence scores and their corresponding observed frequencies. A lower ECE indicates better-calibrated predictions, i.e., the estimated confidence scores are closer to the ground truth labels.

\begin{figure}[htbp]
    \centering
    \includegraphics[width=\textwidth]{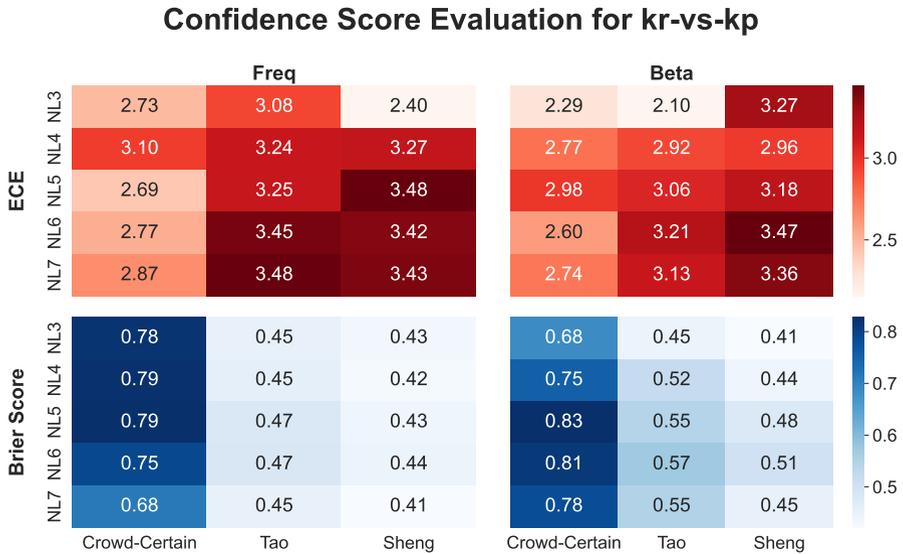}
    \caption[ECE and Brier Score Comparison across Techniques for Varying Number of Workers for kr-vs-kp Dataset] {Comparative evaluation of the ECE and the Brier Score across two confidence score measurement strategies: Beta ($F_{\beta}$) and Freq ($F_{\Omega}$). The results are plotted for three distinct label aggregation techniques --- Crowd-Certain, Tao, and Sheng --— on the kr-vs-kp dataset with varying numbers of crowd workers from three to seven (NL3 to NL7). }\label{fig:crowd.Fig.confidence_score_one_datasets_all_workers}
\end{figure}

Brier Score is a metric that quantifies the accuracy of probabilistic predictions. It calculates the mean squared difference between the estimated confidence scores ($F_{\Omega}^{(i,k)}$ and $F_{\beta}^{(i,k)}$) and the ground truth labels ($y^{(i,k)}$). Hence, higher Brier Score values correspond to better model performance. In Figure~\ref{fig:crowd.Fig.confidence_scores_all_datasets_3_workers}, it can be observed that for the Brier Score metric for both Beta ($F_{\beta}$) and Freq ($F_{\Omega}$) strategies, across all datasets, the proposed Crowd-Certain strategy consistently achieves higher scores when compared to Tao and Sheng. This indicates that the Crowd-Certain strategy offers better-calibrated predictions, providing a higher level of confidence in the aggregated labels.
For the ECE metric and Beta strategy ($F_{\beta}$) the Crowd-Certain strategy outperforms Tao and Sheng across most datasets. For the ECE metric and Freq strategy ($F_{\Omega}$), the Tao technique generally results in higher ECE, indicating worse calibration, whereas the Crowd-Certain and Sheng techniques show varying performance depending on the number of workers.
Figure~\ref{fig:crowd.Fig.confidence_score_one_datasets_all_workers} showcases the results for two metrics, ECE and Brier Score, for two confidence measurement techniques (Beta ($F_{\beta}$) and Freq ($F_{\Omega}$) strategies), applied using three different techniques: Crowd-Certain, Tao, and Sheng. These results are obtained for the kr-vs-kp dataset under different numbers of workers from 3 (denoted with NL3) up to (denoted with NL7). In general, the Brier Score decreases and ECE increases as the number of workers increases, which suggests that increasing the number of workers does not necessarily improve the performance. The performance varies depending on the confidence measurement technique and the strategy used. For the Freq strategy, the Crowd-Certain technique yields lower ECE and higher Brier Score across nearly all numbers of workers compared to the Tao and Sheng techniques, indicating better calibrated predictions when having only 3 workers. For the Beta strategy, the performance varies between techniques. For the Brier Score, the Freq strategy combined with the Crowd-Certain technique performs better across all numbers of workers compared to other combinations of techniques and strategies. For the ECE, the Beta strategy combined with the Crowd-Certain technique yields the lowest values for three and four workers, indicating a good match between predicted confidences and observed frequencies. However, the ECE generally exhibits a tendency to increase as the number of workers increases, indicating a decline in calibration.
Overall, these results suggest that the choice of the confidence measurement technique and the strategy have significant impacts on the calibration (confidence) of the predictions. Further investigations could be beneficial to understand the specific conditions under which certain techniques and strategies yield superior performance.
\section{Discussion}\label{sec:crowd.discussion}
Label aggregation is a critical component of crowdsourcing and ensemble learning strategies. Many generic label aggregation algorithms fall short because they do not account for the varying reliability of the workers. In this work, we introduced a new method for crowd labeling aggregation termed as Crowd-Certain. This technique effectively leverages uncertainty measurements to refine the aggregation of labels obtained from multiple workers. Through an extensive comparative analysis, it was shown to yield higher accuracy in label aggregation against ground truth, particularly in settings where only a limited number of workers are available. This advantage over established methods such as Gold Majority Vote, MV, MMSR, Wawa, Zero-Based Skill, GLAD, and Dawid Skene demonstrates the potential of the proposed method in enhancing the reliability of label aggregation in crowdsourcing and ensemble learning applications.
Our approach is distinguished by its application of a weighted soft majority voting scheme, where the weights are determined based on the level of uncertainty associated with each worker's labels. Importantly, the proposed technique takes into account the possibility of consistently inaccurate workers and includes measures to penalize them (shown in Equation~\ref{eq:crowd.eq.10.consistency-penalized}), thus ensuring the credibility of the computed weights $\omega_{\alpha}^{(k)}$. The calculated weights follow a pre-set ground-truth accuracy closely, highlighting the effectiveness of the technique in capturing the quality of workers' labels. Moreover, the Crowd-Certain technique demonstrates an appreciable capability to generate confidence scores that accompany each aggregated label, offering an extended context that can be invaluable in practical applications.
In this study, we evaluated various techniques for aggregating crowdsourced labels and measuring the confidence scores associated with these labels. This evaluation involved two key metrics (ECE and Brier Score) for the evaluation of confidence scores, as well as three metrics (accuracy, AUC, and F1 score) for the evaluation of the aggregated labels ($\nu$). These metrics assessed different facets of model performance: calibration of the confidence scores (how confident the predictions are), and the performance of the aggregated labels against the ground truth. By comparison to existing methodologies, our method demonstrates superior performance across a variety of datasets, yielding higher average accuracy rates. Furthermore, our experiments, which involved varying the number of workers, demonstrated that Crowd-Certain outperforms the benchmark methods in nearly all scenarios, irrespective of the number of workers involved.
Significantly, our technique introduces an advantageous property by assigning a single weight ($\omega_{\alpha}^{(k)}$ for class $k$) to each worker $\alpha$ for all instances in the dataset. Moreover, the application of predicted probabilities ($\eta_{\alpha}^{(i,k)}$) in our method allows for the reuse of trained classifiers on future sample data, which eliminates the need for recurrent simulation processes. This presents a distinct advantage over conventional techniques, which require computationally expensive and time-consuming repeated simulations for every new dataset. It's worth noting that Crowd-Certain outperforms in nearly all evaluated scenarios across the tested datasets with one exception where the accuracy is lower than MMSR technique on tic-tac-toe dataset as shown in Figures~\ref{fig:crowd.accuracy_per_worker} and~\ref{fig:crowd.Fig.accuracy_all_datasets_workers}. This consistency is evident even when considering variance in dataset characteristics, such as `kr-vs-kp', `mushroom', and `spambase'.
In addition to label aggregation, the evaluation of confidence score measurements revealed further advantages of the Crowd-Certain method. When analyzing two confidence score measurement techniques, Freq and Beta, we found that our strategy achieves lower ECE scores compared to Tao and Sheng for most datasets. This implies that Crowd-Certain provides better-calibrated predictions, offering a higher level of confidence in the aggregated labels. Furthermore, Crowd-Certain also outperformed other techniques in terms of Brier Score across all datasets, indicating a higher accuracy of probabilistic predictions.
Our results indicate that the choice of aggregation and confidence measurement technique can significantly impact the performance. Furthermore, it shows that increasing the number of workers does not necessarily improve the performance, as indicated by the general increase in ECE and decrease (for Freq strategy) in Brier Score with a higher number of workers. This suggests a trade-off between the number of workers and the performance, and that the optimal number may depend on the specific context and the chosen techniques.
\section{Conclusion}\label{sec:crowd.conclusion}
The proposed Crowd-Certain label aggregation technique offers a promising solution for crowdsourced labeling tasks by providing a superior accuracy across various settings. Furthermore, it improves computational efficiency by allowing for the reuse of trained classifiers on future sample data, making it a viable option for large-scale data labeling tasks. While our findings are encouraging, further research and validation across more diverse datasets and real-world scenarios are warranted to further refine and enhance this approach. Future work could delve deeper into understanding why certain techniques and strategies outperform others under specific conditions. Further investigations could explore the effects of other factors such as the complexity of the task and the diversity of the crowd, which may impact the performance of different techniques and strategies. Our findings could guide future research and applications in this domain, with potential implications for various fields that rely on crowdsourced data, including machine learning, data science, and citizen science.
\section{Availability of Data and Materials}
The source code can be found at \href{https://github.com/artinmajdi/crowd-certain}{GitHub: @artinmajdi/crowdcertain}
\section*{Competing Interests}
The authors declare that they have no competing interests.
\bibliographystyle{bst/sn-aps}
\bibliography{Better_BibTeX_Zotero.bib}
\end{document}